\documentclass[onecolumn]{IEEEtran}
\pdfoutput=1
\usepackage{url}
\usepackage{color}
\usepackage{tabulary}

\usepackage{amsfonts}
\usepackage{amssymb}
\usepackage{graphicx}

\begin{document}
%
\title{Day-Ahead Hourly Forecasting of Power Generation from Photovoltaic Plants}

\author{Lorenzo~Gigoni,
        Alessandro~Betti,
				Emanuele~Crisostomi,
        Alessandro~Franco,
				Mauro~Tucci,
				\\
        Fabrizio~Bizzarri,
				Debora~Mucci
\thanks{L. Gigoni and A. Betti are with i-EM S.r.l, Via Lampredi 45, 57121 Livorno, Italy.}
\thanks{E. Crisostomi, A. Franco and M. Tucci are with the Department of Energy, Systems, Territory and Constructions Engineering, University of Pisa, Pisa, Italy. Email: emanuele.crisostomi@unipi.it.}
\thanks{F. Bizzarri is with Enel Green Power S.p.A.}
\thanks{D. Mucci is with Renewable Energy Management, Enel Green Power S.p.A.}}


\maketitle

\bibliographystyle{unsrt}

\begin{abstract}
The ability to accurately forecast power generation from renewable sources is nowadays recognised as a fundamental skill to improve the operation of power systems. Despite the general interest of the power community in this topic, it is not always simple to compare different forecasting methodologies, and infer the impact of single components in providing accurate predictions. In this paper we extensively compare simple forecasting methodologies with more sophisticated ones over 32 photovoltaic plants of different size and technology over a whole year. Also, we try to evaluate the impact of weather conditions and weather forecasts on the prediction of PV power generation.
\end{abstract}



%

\IEEEpeerreviewmaketitle
\IEEEoverridecommandlockouts

\section{Introduction}
High penetration levels of Distributed Energy Resources (DERs), typically based on renewable generation, introduce several challenges in power system operation, due to the intrinsic intermittent and uncertain nature of such DERs. In this context, it is fundamental to develop the ability to accurately forecast energy production from renewable sources, like solar photovoltaic (PV), wind power and river hydro, to obtain short- and mid-term forecasts. Accurate forecasts provide a number of significant benefits, namely:
\begin{itemize}
\item
\textbf{Dispatchability: }secure power systems' daily operation mainly relies upon day-ahead dispatches of power plants \cite{Marinelli2014}. Accordingly, meaningful day-ahead plans can be performed only if accurate day-ahead predictions of power generation from renewable sources, together with reliable predictions of the day-ahead load consumption forecasts (e.g., see \cite{Tucci2016}) are available;
\item
\textbf{Efficiency: }as output power fluctuations from intermittent sources may cause frequency and voltage fluctuations in the system (see \cite{Yona2013}), some countries have introduced penalties for power generators that fail to accurately predict their power generation for the next day; thus, some energy producers prefer to underestimate their day-ahead power generation forecasts to avoid to incur in penalties in the next day. Such induced conservative behaviours are clearly not efficient;
\item
\textbf{Monitoring: }mismatches between power forecasts and the actually generated power may be also used by energy producers to monitor the plant operation, to evaluate the natural degradation of the efficiency of the plant due to the aging of some components (see \cite{Bizzarri2013}) or for early detection of incipient faults.
\end{itemize}

\noindent For the previous reasons, the topic of renewable energy forecasting has been also object of some recent textbooks like \cite{Morales2014} and \cite{Kariniotakis2017} that provide overviews of the state-of-the-art of the most recent technologies and applications of renewable energy forecasting. In this context, the objective of this paper is to compare different methodologies to predict day-ahead hourly power generation from PV power plants.

\subsection{State of the Art}

Power generation from PV plants mostly depends on some meteorological variables like irradiance, temperature, humidity or cloud amount. For this reason, weather forecasts are a common input to forecasting methodologies for PV generation. Depending on the specific problem at hand, forecasts may be also necessary at different spatial and temporal scales, as from high temporal resolutions (i.e., of the order of minutes) and very localized (e.g., off-shore wind farms) to coarser temporal resolutions (e.g., hours) and covering an extended geographical area (e.g., a region or a country) for aggregated day-ahead power dispatching problems. At the same time, very different approaches and methodologies have been explored in the literature, based on statistical, mathematical, physical, machine learning or hybrid (i.e., a mix of the previous) approaches. For example, \cite{Yona2013} uses fuzzy theory to predict insolation from data regarding humidity and cloud amount, and then uses Recurrent Neural Networks (RNNs) to forecast PV power generation. Autoregressive (ARX) methods are used in \cite{Yang2015} for short-term forecasts (minute-ahead up to two hour-ahead predictions) using spatio-temporal solar irradiance forecast models. A forecasting model for solar irradiance for PV applications is also proposed in \cite{Shah2015}. The presence of particulate matter in the atmosphere (denoted as Aerosol Index (AI)) is used in \cite{Liu2015} to support an artificial neural network (ANN) to forecast PV power generation.\\
\newline
As for the specific day-ahead hourly forecasting PV power problem, \cite{Larson2016} use add a least-square optimization of Numerical Weather Prediction (NWP) to a simple persistence model, to forecast solar power output for two PV plants in the American Southwest. A multilayer perceptron was used in \cite{Ehsan2016} to predict the power output of a grid-connected 20-kW solar power plant in India. A stochastic ANN was adopted in combination with a deterministic Clear Sky Solar Radiation Model (CSRM) to predict the power output of four PV plants in Italy. A weather-based hybrid method was used in \cite{Yang2014} as well, where a self-organizing map (SOM), a learning vector quantization (LVQ) network, a Support Vector Regression (SVR) method and a fuzzy inference approach were combined together to predict power generation for a single PV plant. In \cite{Li2015} Extreme Learning Machines (ELMs) are used to predict the power generation of a PV experiment system in Shanghai. Finally, we refer the interested readers to the two recent papers \cite{Raza2016, Antonanzas2016}, and to the references therein, for an extensive review of the literature.\\
\newline
A Global Energy Forecasting Competition (GEFCom2014) has recently allowed different algorithms to be compared, in a competitive way, to solve probabilistic energy forecasting problems, see \cite{Hong2016} for a detailed description of the outcome of the competition. GEFCom2014 consisted of four tracks on load, price, wind and solar forecasting. In the last case, similarly to this paper, the objective was to predict solar power generation on a rolling basis for 24 hour ahead, for three solar power plants located in a certain region of Australia (the exact location of the solar power plants had not been disclosed to the participants of the competition) \cite{Hong2016}. An interesting result of the competition was that all the approaches that eventually ranked at the first places of the competition were nonparametric, and actually consisted of a wise combination of different techniques.

\subsection{Contributions}
The performance of the various forecast models are affected by many elements of uncertainties, and in the opinion of the authors it is not always clear how single choices (e.g., the choice of a specific prediction methodology over another) or different factors (e.g., meteorological forecasting errors) contribute to the final prediction error (i.e., in terms of predicted vs. actual power generation). In fact, most of the existing related papers, including the previous references, generally propose a single methodology to perform the power forecasting task, and compare their results with other very basic algorithms, while comparisons among different more sophisticated approaches can not be easily done. In fact, different authors have generally worked on different data-sets, and the final results can not be compared, as these depend on the specific period of the year where the forecasting error was computed (e.g., in winter time the error is usually lower given that the number of hours of non-zero power generation is lower); the error also depends on the specific country (e.g., it is simpler to predict a sunny day in summer in Italy than in other countries with a more irregular weather); and more in general the use of different error metrics, different weather forecasting tools with different accuracies, different sizes of PV plants, and different technologies or installations (e.g., roof-installed PVs vs. ground PVs) all also contribute to invalidate simple comparisons of different forecasting methodologies on the basis of the final accuracy results alone.\\
\newline
One exception to the previous consideration is provided by the already mentioned Global Energy Forecasting Competition 2014, where different (sophisticated) forecasting algorithms were in fact compared, and ranked, upon the same, publicly available, data-set. However, note that the competition only lasted less than three months, thus not allowing one to validate the final rank over different seasons, and only involved three PV plants. From this perspective, our work extends the results of the competition by further comparing the same algorithms that ranked at the first places of the competition over a longer horizon of time, and over a more variegate set of different PV plants.\\
\newline
Accordingly, more specifically, the contributions of this paper are the following:
\begin{itemize}
\item
We extensively compare four state-of-the-art different ``black-box'' forecasting methodologies upon the same set of data (i.e., k-Nearest Neighbours; Neural Networks; Support Vector Regression; and Quantile Random Forest). Such methods have been tailored to address this specific task of interest. Note that k-Nearest Neighbours, Quantile Random Forests and Support Vector Regression were the building blocks of the algorithms ranked at the first places of the Global Energy Forecasting Competition for solar power forecasting (i.e., they ranked first, second and fifth respectively). We also added Neural Networks in the comparison as it has been frequently used by many other researchers, as from the previous state of the art;
\item
We further compare the predictions results with those obtained with a very simple second-order regressive method. While the previous more sophisticated methodologies usually outperform such a simple method, still the improvement is not as large as one might expect. Accordingly, second-order regressive methods may still be regarded accurate enough for most operations; this simple comparison is usually missing in the literature, thus not making it simple to evaluate the gain obtained by using more sophisticated methods;
\item
As a final term of comparison, we also consider an ensemble of all the previous methodologies that combines the forecasts obtained by the single algorithms. This combined method outperforms the single algorithms alone, thus confirming what has been observed in other examples as well (e.g., \cite{Rokach2010}), and consistently with the results in \cite{Hong2016} as well;
\item
All results are obtained and validated upon real data recorded over 32 different PV plants of different sizes, technologies and characteristics in general, for an overall installed nominal power of about 114 MW, for an horizon of one year. Also, the availability of so many PV plants is also an element that is missing in most comparisons existing in the literature;
\item
In addition, we evaluate how the accuracy of the different algorithms vary under different weather conditions;
\item
Finally, we try to evaluate the component of the error that is solely due to the inaccuracy of the weather forecasts, by comparing the accuracy of the same methodology when the inputs are predicted variables from meteorological forecasts, and when the inputs are the actual measured variables.
\end{itemize}

This paper is organised as follows: Section \ref{Case_study} describes more in detail the case study, and other ingredients required to perform the comparison. Section \ref{Methodologies} briefly describes the forecasting methodologies that we have used to predict power generation from PV plants. The obtained results are provided and discussed in Section \ref{Results}. Finally, in Section \ref{Conclusion} we conclude our paper and outline our current line of research in this topic.

\section{Case study}
\label{Case_study}
\begin{figure}[!t]
\centering
\includegraphics[width=\columnwidth]{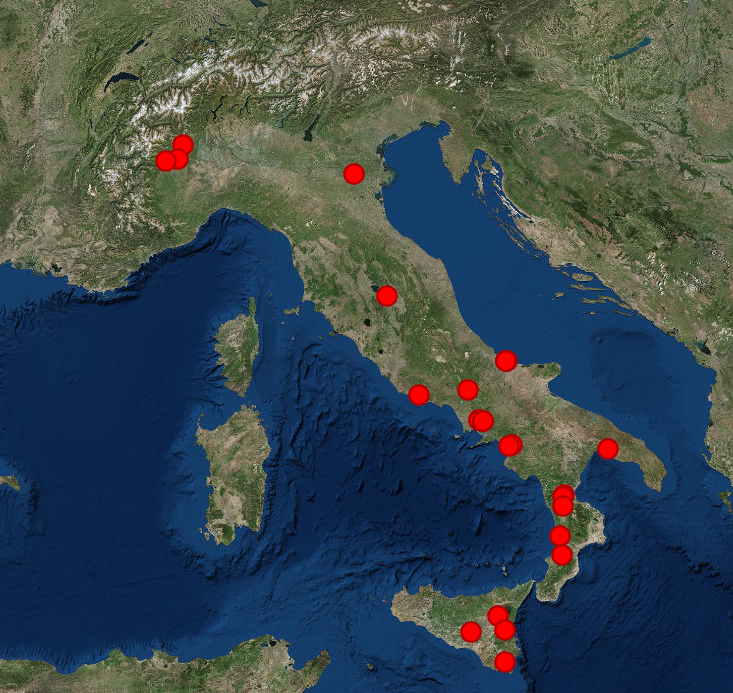}
\caption{Position of the PV plants. Some PV plants are very close to each other and their corresponding circles actually overlap in the figure. As can be seen from the figure, most PV plants are in Southern Italy where more convenient weather conditions for PV plants can be found.}
\label{figura_Ema_2}
\end{figure}

We here use data collected from 32 PV plants installed at different latitudes in Italy (i.e., Northern Italy, Central Italy, Southern Italy and Sicily). The size of the plants ranges from a few tens up to 10 MW, for an overall installed nominal power of about 114 MW. The technology used for the PV films includes monocrystalline silicon, polycrystalline silicon, thin-film amorphous silicon and flexible amorphous thin-film silicon. About half of the PV plants are installed on roofs, and half from the ground (we also had the knowledge of the tilt angles for all PV panels in the PV plants). The investigated set of PV plants was chosen to be a representative (scaled) set of the whole Italian PV installations (both in terms of technological mixture, and geographical positioning). The exact location of the PV plants is shown in Figure \ref{figura_Ema_2}.
For this work, we have used data of the generated power from June 2014 up to end of 2016.

\subsection{Data}
\label{Data}
We can classify the available data as follows:\newline
\\
\textbf{Meteorological data :} We have used meteorological forecasts from two different providers; namely, Aeronautica Militare\footnote{\url{http://www.meteoam.it/}} and MeteoArena\footnote{\url{https://www.meteoarena.com/}}. In particular, we had at our disposal the 24-hour ahead hourly forecast of direct, diffuse, and total irradiance on a horizontal surface on Earth; and the 24-hour ahead hourly forecast of temperature.\\
\newline
\textbf{Measured data :} We had at our disposal the actual hourly power generated from each PV plant. Also, we had the horizontal irradiance measured from satellite data (with a spatial resolution of 3.5 km $\times$ 3.5 km and a time resolution of an hour). We used satellite data to optimally combine the forecasting meteorological data from the two providers into a single value. While we found that such a combination would depend on the specific PV plant (as one meteorological forecasting provider may be more accurate than the other for a specific location), still we do not give here more details of such a combination procedure, as it is not the focus of this paper.\\
\newline
\textbf{Computed data :} We used the knowledge of the time of the day (and the year) to compute the sun altitude and azimuth. Then we used Perez Sky Diffuse Model (see \cite{Perez1990}), together with the knowledge of the data of single PV plants, to infer the beam, the diffuse and the total irradiance upon the tilted plane of the panels.

\subsection{Training set and validation set}
We used hourly data from 1 May 2014 until 8 November 2015 as our training set, and hourly data from 9 November 2015 to 12 November 2016 as a test set (5 weeks were actually not used due to the lack of information from one weather forecasting provider; this difference with respect to the other weeks of the year did not allow us to validate the results in those periods. Accordingly, there are 2 missing weeks in May 2016, and 3 missing weeks in August 2016). This implies that the test set is about one year long (48 weeks), which gives us the possibility to compare the accuracy of PV forecasts over all seasons. Also, every week we extend the training set to include the last available measurements, which is convenient both to increase the size of the training set, and also to take into account the latest effects (e.g., aging of the PV plants).

\subsection{Performance Indices}
\label{Performance_Index}
We use the normalised Mean Absolute Error (nMAE) as the main performance index to compare the different algorithms. In particular, nMAE is defined as
\begin{equation}
\label{NMAE}
nMAE = \displaystyle \frac{1}{N} \displaystyle \sum_{i=1}^{N} \displaystyle \frac{\left|\widehat{P}(i)-P_m(i)\right|}{P_n(i)} \cdot 100,
\end{equation}
where $\widehat{P}(i)$ is the predicted generated power at the $i$'th hour, and $P_m(i)$ and $P_n(i)$ are the actual measured power and the nominal power of that given PV plant respectively, at the same hour (note that the nominal power may actually change over time if part of the PV plant is unavailable, e.g., due to maintenance reasons). The prediction error is then averaged upon the number of hours $N$ of the comparison horizon. The nMAE is frequently used to evaluate forecasting errors, as it allows one to better compare the results obtained for plants of different size (see for instance \cite{DeGiorgi2014}).\\
\newline
Other performance indices are however frequently used in the related literature as well. Among others, in this paper we shall further consider a normalised Root Mean Square Error (nRMSE), the Mean Absolute Error (MAE) and the normalised Mean Bias Error (nMBE) (see for instance \cite{Larson2016} where the same performance indicators have been recommended as well). In particular, the nRMSE penalyses large errors more than the nMAE, and it is defined as
\begin{equation}
\label{RMSE}
nRMSE = \displaystyle \sqrt{\frac{1}{N} \displaystyle \sum_{i=1}^{N} \displaystyle \left(\displaystyle \frac{\widehat{P}(i)-P_m(i)}{P_n(i)}\right)^{2}} \cdot 100.
\end{equation}
Both the nRMSE and the nMAE do not retain the information on the sign of the error. For this reason, the nMBE may be useful, as it is defined as 
\begin{equation}
\label{MBE}
nMBE = \displaystyle \frac{1}{N} \displaystyle \sum_{i=1}^{N} \displaystyle \left(\displaystyle \frac{\widehat{P}(i)-P_m(i)}{P_n(i)}\right) \cdot 100,
\end{equation}
where a positive nMBE corresponds to an overestimation of the actual power generation. It is worth to mention that indices (2) and (3) were recently recommended by the European and International Energy Agency
(IEA) for reporting irradiance model accuracy \cite{IEA2012}. 
Finally, the MAE may be useful as, differently from all the previous indicators does not have normalisation factors at the denominator: 
\begin{equation}
\label{MAE}
MAE = \displaystyle \frac{1}{N} \displaystyle \sum_{i=1}^{N} \displaystyle \left|\widehat{P}(i)-P_m(i)\right|,
\end{equation}

\section{Methodologies}
\label{Methodologies}
\begin{figure}[!t]
\centering
\includegraphics[width=\columnwidth]{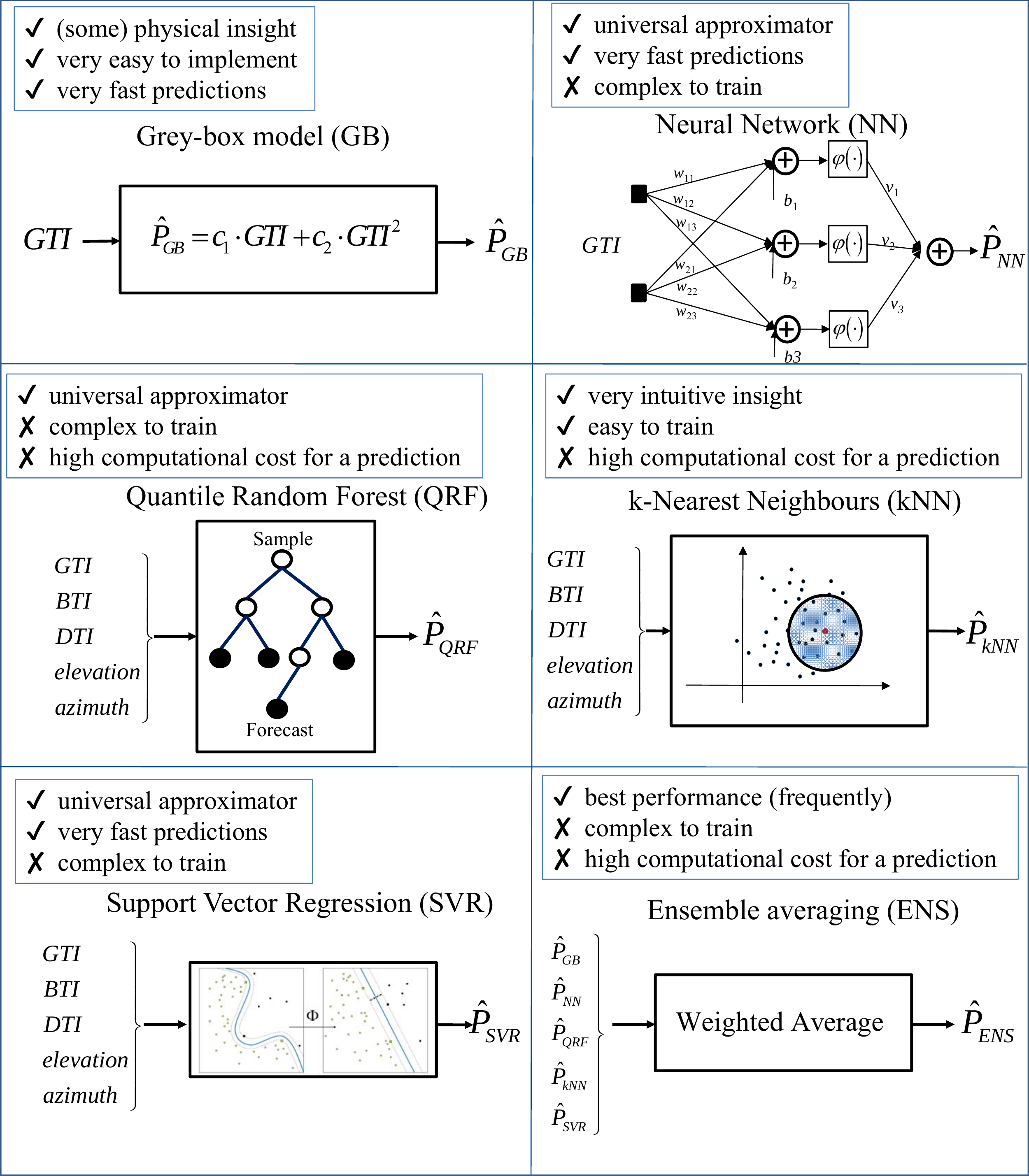}
\caption{Summary of the six compared methodologies. The GB model (top-left) is clearly the simplest method to implement. However, we shall later show that other methodologies, though more complex and computationally intensive, provide more accurate forecasts. Also, the methodology that combines the other five forecasts (bottom right) outperforms the single strategies.}
\label{Different_Methodologies}
\end{figure}

Many different methodologies may be adopted to predict power generation from PV plants. A typical classification of different methodologies relies on whether the physical equations of the PV plants are in fact used or not. For instance, reference \cite{Bizzarri2013} develops a novel electrothermal macro-model of PV power plants that takes as input variables the ambient temperature, irradiance and wind speed, and gives the generated power as an output. In the experience of the authors, such models are convenient in many situations (e.g., for initial plant design), but may not be effective for actual power generation forecast for many reasons: (i) such models are in any case (possibly accurate) \textit{approximations} of the true plant, given the difficulty in modelling every possible physical phenomenon occurring in a complex real-world power plant; (ii) in any case, such models require a perfect real-time knowledge of all physical variables appearing in the power plant, which in most cases are not fully available; (iii) physical variables degrade over time anyway due to aging factors (so one should continuously monitor all variables all the time).\newline
\\
In order to overcome the previous shortcomings of physical models, many authors have adopted so-called ``black-box'' strategies, where the physical relationships between input variables and the generated power have been completely neglected, and let some Machine Learning (ML) methodologies (e.g., neural networks) choose the most convenient (arbitrary) relationship between input and output signals. Note that in this case one loses the physical meaning of the input-output relationship, and that the accuracy of such methods usually relies on the ability to have a large enough database that allows one to construct a reliable model. However, such methods are known to work rather well in practice (see for instance \cite{Yona2013} and \cite{Liu2015} for two specific examples, and \cite{Raza2016, Antonanzas2016} for an overview of all ML algorithms employed in the literature for PV power forecasts). As a further example, all approaches that ranked highly in the GEFCom2014 solar  competition were nonparametric \cite{Hong2016}.\newline
\\
Sometimes, reasonably accurate models can be achieved also if so-called ``grey-box'' identification methods are used. In this case, the input-output relationship is imposed from the outside, using some simplified physical relations (e.g., by neglecting high-order dependencies). As we shall see in the next section, such simple models may also provide accurate predictions.\newline
\\
The specific methodologies are now described in greater detail, while a schematic summary is provided in Figure \ref{Different_Methodologies}.

\subsection{Grey-box model (GB)}
\label{Grey-box model}
We use here a very simple second-order model
\begin{equation}
\label{GB_Equation}
\widehat{P}_{GB}(i) = c_1 \cdot GTI(i) + c_2 \cdot GTI(i)^2 + c_3 \cdot GTI(i) \cdot T(i),
\end{equation}
where $\widehat{P}_{GB}(i)$ is the predicted hourly generated power, using the Grey-box model, and $GTI(i)$ is the hourly forecast of the Global Irradiance, projected on the tilted plane of the panel. This simple model follows from the popular PVUSA model (see \cite{PVUSA}) that is a simplification of the true relationship between temperature and irradiance as input variables, and generated power as an output. This model is widely used as a starting point for PV forecasting problems (see \cite{Bianchini2013} for example). Parameters $c_1 > 0, c_2 < 0$ and $c_3 < 0$ are three model parameters that depend on the specific power plant (e.g., technology, area of the surface of the panels, angles), and are estimated from available historical data. In our specific applications, we found it to be more accurate by enforcing $c_3$ to be zero (i.e., neglecting the mixed term that both depends on the temperature and irradiance), and used a pseudo-inversion to obtain the values of $c_1$ and $c_2$ that best fitted the data. In our experience, we found it to be more convenient to update the parameters $c_1$ and $c_2$ every week, using only data of the last four weeks. The motivation for this is that, most likely, the optimal parameters of this simplified model change according to different climate conditions over the year, thus it is convenient to only consider the most recent history. Also, the conditions of the PV plants may change in time due to several factors (e.g., aging factors).

\subsection{Neural Network (NN)}
\label{NN}
We have used a static feed-forward neural network, with a hidden layer of 3 neurons, a sigmoidal activation function, and used a Bayesian regularization function for training. All choices were performed by trying different combinations on the same training set, and comparing the outcomes upon the validation set. The inputs were the same of the GB algorithm, and again, retraining was performed every week to possibly change the internal weights of the neural network. However, for simplicity, we decided not to change the structure of the neural network (e.g., number of neurons or activation function) during the test period.

\subsection{k-Nearest Neighbours (kNN)}
\label{k-NN}
The k-Nearest Neighbours is one of the simplest methods in machine learning. The rationale of this methodology is that a given weather forecast will most likely give rise to a power generation that will be very close to those in the past when there were similar weather conditions. Accordingly, one searches in the past data-base for $k$ weather forecasts that are the most similar to those of the hour of interest (i.e., the $k$ nearest neighbours in the name of the algorithm). Then, the corresponding historical power generations are combined (e.g., by giving more importance to those corresponding to the most similar weather conditions) to provide a single forecast of power generation. This algorithm has a number of free parameters (i.e., how to normalize the train set; how to compute the distance between the vector of predicted weather variables and a historical vector of the same variables; how many neighbours should be considered; and how to weigh the neighbours to combine the corresponding power generations to provide a single value). In this work, we have normalized variables so that each meteorological variable lies in the $[0, \ 1]$ interval; we have used Euclidean distance to compute the distance between two sets of meteorological variables; we have chosen $k=300$; and have finely used a weighted average to combine the outputs of the $k$ nearest neighbours, with weights chosen according to a Gaussian similarity kernel, i.e., 
\begin{equation}
\label{kNN_Equation}
w_i = e^{\displaystyle -\frac{d_i^2}{\sigma^2 \cdot d_1^2}},
\end{equation}
where the weight of the historical $i$th neighbour $w_i$ is computed as a function of its Euclidean distance $d_i$ from the weather forecast of the hour of interest. In (\ref{kNN_Equation}), $d_1$ is the distance of the closest neighbour (i.e., $d_1=min\left\{d_i\right\},i=1,...,k$ and $\sigma$ is a parameter of the algorithm (in our case, we chose $\sigma=4$). The weights are then normalized to sum to unity. Note that this choice is slightly different from the conventional distance used by k-NN algorithms (that is the hyperbolic one), but is consistent with other similar works (see \cite{Tucci2016}). All parameters were learned over the training set and validated over the validation set.\\
\newline
In this case, we used a larger number of inputs, as in addition to the tilted global irradiance (GTI), we also used the diffuse (DTI) and the beam (BTI) components of the global irradiance, and the azimuth and elevation angles of the sun (the use of such inputs did not apparently give benefits in the case of the NN, but did actually give an improvement of the performance of the k-NN algorithm, in the validation set). In this case, we used the weekly ``retraining'' to increase the size of the historical dataset, but we did not change the parameters of the algorithm.

\subsection{Quantile random forest (QRF)}
\label{RF}
Random forest is a stochastic machine learning algorithm, originally developed in the '90s and later extended by Breiman \cite{Breiman2001}. The random forest algorithm is actually an ensemble of classification models, where each model is a decision tree. The rationale behind this approach is that combining multiple classification models increases predictive performance than having a single decision tree. However, caution is required to build an ensemble of uncorrelated decision trees, and either \textit{boosting} or \textit{bagging} techniques can be used for this objective. In particular, random forest can be seen as a special case of bagging, where a further node splitting stage is added. In this work, we have used 300 trees, 5 (minimum) samples at the terminal nodes (i.e., leaves), Minimum Square Error (MSE) as objective function for the splitting method, and the hourly power as output variable. Finally, we have actually deployed the Quantile Random Forest variant of the algorithm, see \cite{Meinshausen2006}, that, differently from conventional random forests, takes track of all the target samples, and not just of their average. As a further parameter at this regard, we have obtained 0.4 as the optimal value of the quantile parameter. All values have been obtained upon (again) the same validation set.\\
\newline
Again, in the case of the random forest, we have found it convenient to use the same number of inputs of the k-NN. While we have maintained again the same structure of the algorithm (i.e., in terms of number of trees and quantile parameter) during the weekly retraining, yet note that, at least in principle, the retraining causes the determination of a different decision tree every week.

\subsection{Support vector regression (SVR)}
\label{SVM}
Support vector regression methods were originally proposed in \cite{Vapkin1995, Vapkin1997} as an application of Support Vector Machine (SVM) theory to regression estimation problems. One of the main characteristics of SVM techniques is that the so-called \textit{structural risk} is minimised in the training phase, rather than the output training error (e.g., as in NNs). Among the many existing formulations of SVR, here we used $\nu$-SVR \cite{Smola2004}, where the parameter $\nu \in (0,1]$ represents an upper bound on the training error and a lower bound on the fraction of used support vectors. In addition to $\nu$, the algorithm requires also a meta-parameter $\gamma$, that is required to define the Gaussian kernel function, and $C$ that is the regularization parameter.\\
\newline
Again, for SVR we used the same inputs of the k-NN and QRF, and the same validation set, where we obtained $\nu=0.5$, $\gamma=1.25$ and $C=1$ as the optimal values of the parameters, and did not change them during our weekly retraining. 

\subsection{Ensemble of methods (ENS)}
\label{Ensemble}
The last method included in the comparison is the ensemble of all of the previous methods. The main idea behind the ensemble methodology is to weigh several individual independent forecasting techniques, and combine them in order to obtain another forecast that outperforms every one of them \cite{Rokach2010}. In fact, as noticed among others in \cite{Rokach2010} again, human beings tend to seek several opinions before making any important decision.\\
\newline
While different ways can be devised to combined single forecasts, here we have used the so-called stacked generalization \cite{Wolpert1992}, that simply computes a weighted average of the outputs of the previous algorithms. The optimal weights were here simply computed by minimising the square error in the validation set between the powers predicted by every single method, and the actual generated powers (i.e., using the Moore–Penrose pseudoinverse). Then we normalised the weights to have a unit sum.\\
\newline
While this was a simple way to combine the different forecasts, still it eventually outperformed all the other techniques singularly adopted, as it will be described in the next section.

\section{Experimental results}
\label{Results}
\begin{figure*}[!t]
\centering
\includegraphics[width=\textwidth]{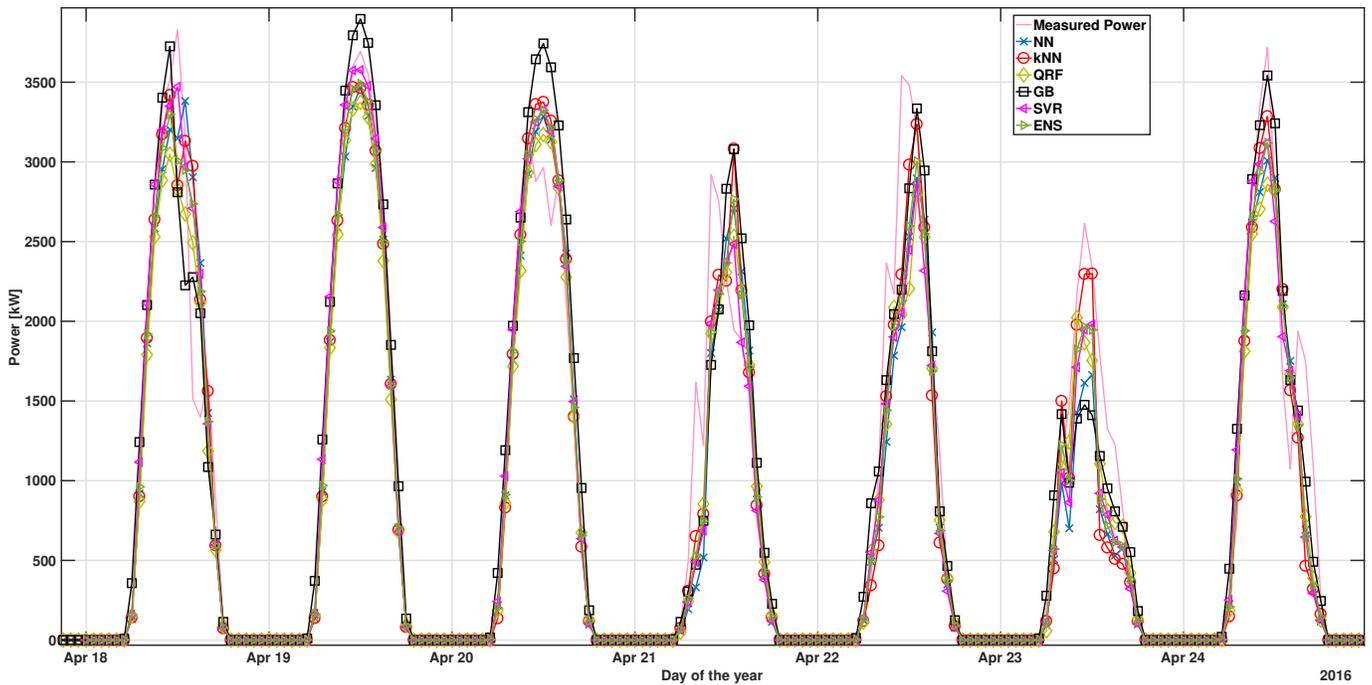}
\caption{Generated power vs. predicted power in seven following days in April 2016.}
\label{Weekly_Behaviour}
\end{figure*}

\begin{figure*}[!t]
\centering
\includegraphics[width=\columnwidth]{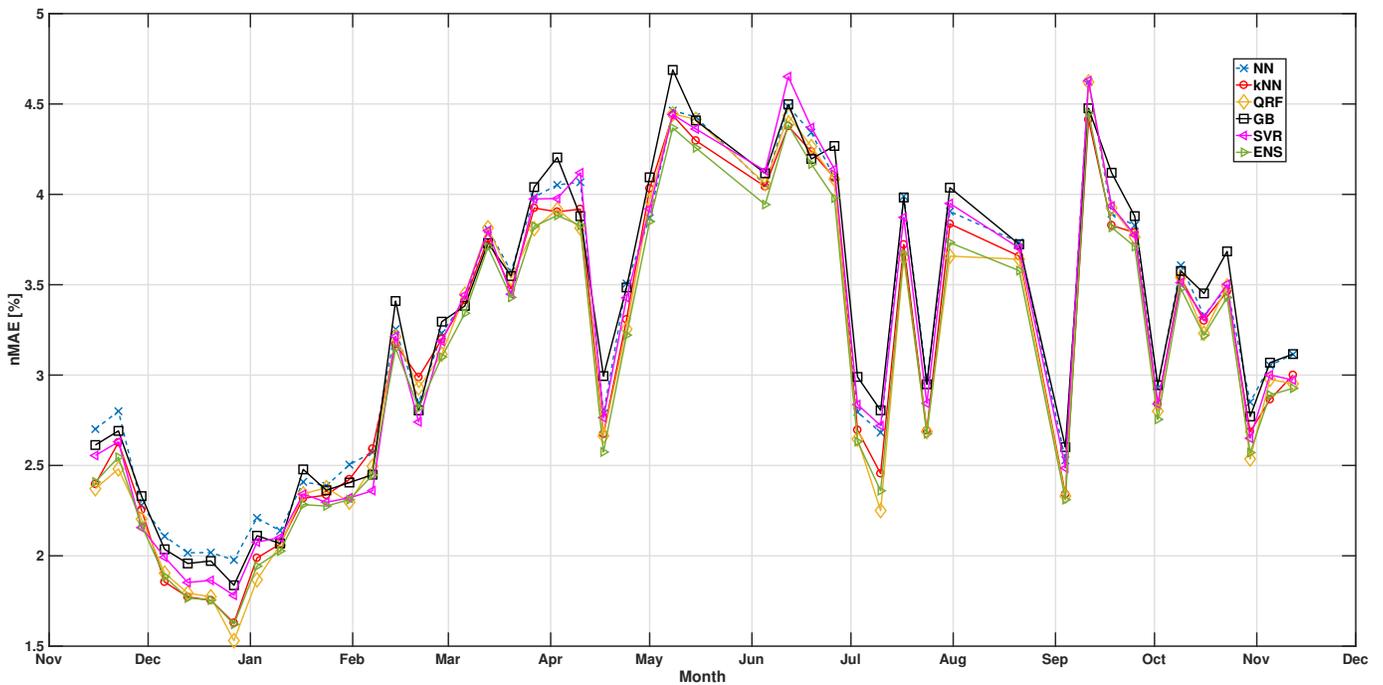}
\caption{Weekly error (nMAE) obtained by each forecasting methodology. The range of data is from November 2015 to November 2016.}
\label{Weekly_Error}
\end{figure*}

Figure \ref{Weekly_Behaviour} compares the generated (measured) power versus the predicted powers, obtained with the six compared forecasting methodologies, for a PV plant of nominal size of 4834 kW (i.e., PV plant number 17 in Table \ref{PV_Plant_Comparison}). The comparison was performed in seven following days in April 2016. Both the time range and the specific plant were here randomly chosen just for the purpose of this figure. In particular, the figure shows the typical characteristic bell-shaped curves of solar power generation, and the generated power is practically zero at night time. 
More detailed results of the comparison are given in Figure \ref{Weekly_Error}, that shows the weekly nMAE error (computed according to Equation (\ref{NMAE}), and aggregated and averaged over an interval of a week for all 32 plants) obtained by each forecasting methodology. 
Two main interesting aspects can be noted from Figure \ref{Weekly_Error}. The first one is that the errors depend on the particular season of the year. In particular, errors are larger in May and September, when solar power generation in Italy is still relatively high, but the weather is more unpredictable than in summer months. A second observation is also a very simple model like the GB is generally accurate at predicting the correct power generation shape. Indeed, its prediction rather overlaps those of the other methodologies. This shows that a second-order approximation of the true physics underlying the solar power generation problem may already provide an accurate forecast, if the method is tuned in a careful way. Still, other more complex methodologies do outperform GB, and are more convenient when a better performance is required.\\
\newline
A more detailed information regarding the comparison, together with other evaluation metrics, are shown in Table \ref{Monthly_Comparison}. From Table \ref{Monthly_Comparison} it is possible to appreciate that the combination of single methodologies (ENS) outperforms the single techniques. Also, ENS, QRF and kNN usually provide the three best performance, while SVR, NN and GB provide in general less accurate forecasts. This result is aligned with that of the GEFCom2014 competition (\cite{Hong2016}) where QRF and kNN had outperformed all other methods (including SVR). Also, the fact that ENS is the best performing algorithm confirms the theoretical results of \cite{Rokach2010} and is also consistent with the experimental results of other works, see for instance \cite{Alessandrini2015, Ren2015}, indicating that the combination of several different forecasting tools is the best approach for the prediction of solar power generation.\\
\newline
The index nMBE deserves a special discussion: this index evaluates the signed bias of the algorithms. Thus, a small value of nMBE does not imply an accurate forecast (but rather, that errors compensate each other). While all algorithms have a small value of the nMBE, still it is possible to see that all the algorithms eventually overestimate power generation (between $0.156\%$ and $0.514\%$). Also note that ENS does not provide the best performance in terms of nMBE as it provides an intermediate forecast that tends to mitigate the error dispersion, but not its error bias. 

\begin{table*}[tb!]
	\caption{Monthly error.} \centering
	
	\resizebox{\linewidth}{!}{
	\begin{tabular}{|l|l||l|l|l|l|l|l|l|l|l|l|l|l||l|}
		\hline
		\multicolumn{2}{|c||}{Error-Method / Month} & \multicolumn{1}{|c|}{Jan}& \multicolumn{1}{|c|}{Feb} & \multicolumn{1}{|c|}{Mar} & \multicolumn{1}{|c|}{Apr}& \multicolumn{1}{|c|}{May} & \multicolumn{1}{|c|}{Jun} & \multicolumn{1}{|c|}{Jul} & \multicolumn{1}{|c|}{Aug} & \multicolumn{1}{|c|}{Sep} & \multicolumn{1}{|c|}{Oct}& \multicolumn{1}{|c|}{Nov} & \multicolumn{1}{|c||}{Dec} & Avg.\\
		\hline
		\hline
		& GB & 2.35 & 3.00 & 3.62 & 3.79 & 4.45 & 4.21 & 3.35 & 3.40 & 3.68 & 3.39 & 2.72 & 1.93 & 3.26\\
		\cline{2-15}
		& NN & 2.39 & 2.99 & 3.61 & 3.73 & 4.35 & 4.14 & 3.28 & 3.42 & 3.64 & 3.34 & 2.75 & 2.01 & 3.24\\
		\cline{2-15}
nMAE & kNN & 2.32 & 3.00 & 3.55 & 3.64 & 4.30 & 4.06 & 3.08 & 3.28 & 3.53 & 3.26 & 2.58 & 1.73 & 3.13\\
		\cline{2-15}
(\%) & QRF & 2.29 & 2.95 & 3.57 & 3.58 & 4.35 & 4.09 & \textbf{2.97} & 3.26 & 3.58 & 3.22 & 2.55 & \textbf{1.72} & 3.11\\
		\cline{2-15}
		& SVR & 2.30 & 2.90 & 3.56 & 3.73 & 4.34 & 4.20 & 3.25 & 3.37 & 3.61 & 3.27 & 2.62 & 1.85 & 3.19\\
		\cline{2-15}
		& ENS & \textbf{2.25} & \textbf{2.90} & \textbf{3.49} & \textbf{3.54} & \textbf{4.23} & \textbf{3.99} & 3.02 & \textbf{3.22} & \textbf{3.49} & \textbf{3.20} & \textbf{2.54} & 1.73 & \textbf{3.07}\\
 	\hline
	\hline
		& GB & 5.97 & 7.07 & 7.68 & 7.82 & 9.05 & 8.35 & 7.92 & \textbf{8.43} & 7.91 & 7.37 & 6.64 & 4.91 & 7.36\\
		\cline{2-15}
		& NN & 5.94 & 7.00 & 7.65 & 8.01 & 9.19 & 8.44 & 7.97 & 8.62 & 7.92 & 7.30 & 6.64 & 5.18 & 7.41\\
		\cline{2-15}
nRMSE & kNN & 5.80 & 6.86 & \textbf{7.45} & 7.72 & 8.94 & 8.27 & 7.74 & 8.55 & 7.66 & 7.11 & 6.49 & 4.85 & 7.21\\
		\cline{2-15}
(\%) & QRF & 5.93 & 6.93 & 7.73 & 7.80 & 9.15 & 8.43 & 7.66 & 8.57 & 7.83 & 7.22 & 6.59 & 4.89 & 7.32\\
		\cline{2-15}
		& SVR & 5.98 & 6.93 & 7.73 & 8.04 & 9.09 & 8.47 & 7.86 & 8.58 & 7.94 & 7.30 & 6.67 & 5.06 & 7.40\\
		\cline{2-15}
		& ENS & \textbf{5.76} & \textbf{6.78} & 7.46 & \textbf{7.67} & \textbf{8.88} & \textbf{8.19} & \textbf{7.65} & 8.45 & \textbf{7.66} & \textbf{7.08} & \textbf{6.45} & \textbf{4.78} & \textbf{7.16}\\
		\hline
		\hline
		& GB & 102 & 128 & 136 & 140 & 154 & 148 & 114 & 129 & 142 & 133 & 115 & 84.8 & 126\\
		\cline{2-15}
		& NN & 104 & 128 & 134 & 135 & 149 & 146 & 115 & 130 & 140 & 132 & 116 & 87.7 & 125\\
		\cline{2-15}
MAE	& kNN & 102 & 128 & 134 & 132 & 146 & 140 & 108 & 125 & 137 & 127 & 108 & \textbf{73.4} & 120\\
		\cline{2-15}
(kW) & QRF & 102 & 126 & 136 & 130 & 149 & 139 & \textbf{103} & 124 & 140 & 127 & 107 & 73.7 & 120\\
		\cline{2-15}
		& SVR & 101 & 124 & 134 & 135 & 148 & 146 & 116 & 128 & 140 & 130 & 110 & 79.1 & 123\\
		\cline{2-15}
		& ENS & \textbf{99.5} & \textbf{124} & \textbf{132} & \textbf{128} & \textbf{144} & \textbf{138} & 105 & \textbf{122} & \textbf{136} & \textbf{126} & \textbf{107} & 74.1 & \textbf{118}\\
		\hline
		\hline
		& GB & 0.226 & 0.349 & 0.255 & \textbf{0.525} & \textbf{0.150} & 0.538 & -0.0867 & 1.29 & -0.396 & 0.527 & 0.545 & 0.398 & 0.313\\
		\cline{2-15}
		& NN & 0.431 & 0.422 & 0.385 & 1.31 & 0.675 & 0.658 & 0.154 & 1.60 & \textbf{-0.187} & 0.324 & 0.627 & 0.565 & 0.514\\
		\cline{2-15}
nMBE	& kNN & 0.487 & 0.429 & 0.367 & 1.03 & 0.545 & 0.313 & -0.0841 & \textbf{1.29} & -0.248 & 0.295 & 0.766 & 0.607 & 0.434\\
		\cline{2-15}
(\%) & QRF & \textbf{0.0912} & \textbf{-0.0406} & \textbf{-0.0874} & 0.751 & 0.313 & \textbf{0.296} & -0.189 & 1.39 & -0.342 & \textbf{-0.0519} & \textbf{0.377} & \textbf{0.238} & \textbf{0.156}\\
		\cline{2-15}
		& SVR & 0.504 & 0.380 & 0.436 & 1.41 & 0.728 & 0.381 & \textbf{-0.0712} & 1.52 & -0.279 & 0.255 & 0.697 & 0.720 & 0.493\\
		\cline{2-15}
		& ENS & 0.334 & 0.256 & 0.224 & 1.02 & 0.495 & 0.382 & -0.0848 & 1.41 & -0.287 & 0.191 & 0.589 & 0.487 & 0.356\\
 	\hline
	\end{tabular}
	}
	\label{Monthly_Comparison}
\end{table*}
Similarly, Table \ref{PV_Plant_Comparison} compares the six forecasting algorithms over every single PV plant, where the nominal sizes of the plants are reported as well. The results in Table \ref{PV_Plant_Comparison} confirm that ENS consistently outperforms the other algorithms (with the three plants being an exception). This further confirms that despite the differences are not huge, still they are consistently true.
\begin{table}[tb!]
	\caption{Average error (nMAE), plant by plant.} \centering
	
	\resizebox{\linewidth}{!}{
	\begin{tabular}{|l|l|l|l|l|l|l|l|}
		\hline
		\multicolumn{1}{|r|}{PV \#} & \multicolumn{1}{|c|}{$P_{nom}$ [kW]} & \multicolumn{1}{|c|}{GB}& \multicolumn{1}{|c|}{NN} & \multicolumn{1}{|c|}{kNN} & \multicolumn{1}{|c|}{QRF} & \multicolumn{1}{|c|}{SVR} & \multicolumn{1}{|c|}{ENS}\\
		\hline
		1 & 555 & 3.06 & 3.04 & 3.00 & 3.07 & 3.12 & \textbf{2.98}\\
		\hline
		2 & 960 & 2.97 & 2.90 & 2.93 & 2.97 & 2.99 & \textbf{2.89}\\
		\hline
		3 & 9950 & 4.23 & 4.17 & 3.98 & 3.96 & 4.13 & \textbf{3.93}\\
		\hline
		4 & 9950 & 4.14 & 4.38 & 4.16 & 4.05 & 4.28 & \textbf{4.04}\\
		\hline
		5 & 4170 & 1.92 & 1.78 & 1.76 & 1.73 & 1.77 & \textbf{1.72}\\
		\hline
		6 & 3750 & 1.30 & 1.33 & 1.29 & 1.29 & 1.36 & \textbf{1.27}\\
		\hline
		7 & 5250 & 3.02 & 3.10 & 2.97 & 2.93 & 2.98 & \textbf{2.91}\\
		\hline
		8 & 3050 & 3.18 & 3.65 & 3.34 & \textbf{3.04} & 3.38 & 3.17\\
		\hline
		9 & 2260 & 2.89 & 2.84 & 2.79 & 2.77 & 2.85 & \textbf{2.72}\\
		\hline
		10 & 1760 & 2.90 & 2.71 & 2.70 & 2.75 & 2.75 & \textbf{2.65}\\
		\hline
		11 & 4991 & 4.00 & 3.82 & 3.58 & 3.57 & 3.65 & \textbf{3.56}\\
		\hline
		12 & 1935 & 3.98 & 3.92 & 3.82 & 3.77 & 3.89 & \textbf{3.75}\\
		\hline
		13 & 1070 & 3.36 & 3.32 & 3.25 & 3.21 & 3.33 & \textbf{3.16}\\
		\hline
		14 & 2150 & 1.95 & 1.87 & 1.78 & \textbf{1.72} & 1.76 & 1.73\\
		\hline
		15 & 8723 & 3.71 & 3.66 & 3.57 & 3.71 & 3.61 & \textbf{3.55}\\
		\hline
		16 & 1190 & 1.82 & 1.98 & 1.81 & 1.79 & 1.96 & \textbf{1.78}\\
		\hline
		17 & 4834 & 3.57 & 3.63 & 3.40 & 3.35 & 3.48 & \textbf{3.30}\\
		\hline
		18 & 6000 & 3.84 & 3.84 & 3.66 & 3.56 & 3.71 & \textbf{3.56}\\
		\hline
		19 & 6600 & 3.90 & 3.87 & 3.74 & 3.73 & 3.86 & \textbf{3.67}\\
		\hline
		20 & 4540 & 3.17 & 3.05 & 2.98 & 3.01 & 3.00 & \textbf{2.93}\\
		\hline
		21 & 1560 & 3.94 & 3.85 & 3.64 & 3.65 & 3.75 & \textbf{3.63}\\
		\hline
		22 & 9990 & 3.92 & 3.86 & 3.75 & 3.81 & 3.82 & \textbf{3.72}\\
		\hline
		23 & 1410 & 4.20 & 4.25 & 4.10 & 4.06 & 4.07 & \textbf{4.04}\\
		\hline
		24 & 1200 & 3.83 & 3.73 & 3.61 & \textbf{3.56} & 3.70 & 3.57\\
		\hline
		25 & 2630 & 2.94 & 3.02 & 2.94 & 2.95 & 2.98 & \textbf{2.89}\\
		\hline
		26 & 2010 & 3.53 & 3.62 & 3.48 & 3.52 & 3.48 & \textbf{3.41}\\
		\hline
		27 & 1000 & 3.78 & 3.84 & 3.70 & 3.63 & 3.72 & \textbf{3.61}\\
		\hline
		28 & 8240 & 4.00 & 3.77 & 3.63 & 3.67 & 3.75 & \textbf{3.61}\\
		\hline
		29 & 480 & 2.82 & 2.75 & 2.71 & 2.66 & 2.75 & \textbf{2.63}\\
		\hline
		30 & 720 & 2.76 & 2.67 & 2.65 & 2.59 & 2.64 & \textbf{2.56}\\
		\hline
		31 & 480 & 2.91 & 2.84 & 2.81 & 2.77 & 2.82 & \textbf{2.73}\\
		\hline
		32 & 660 & 2.79 & 2.75 & 2.73 & 2.75 & 2.72 & \textbf{2.66}\\
		\hline					
	\end{tabular}
	}
	\label{PV_Plant_Comparison}
\end{table}
      
\begin{figure*}[ht!]
\resizebox{\linewidth}{!}{
 \begin{tabular}{cc}
    \includegraphics[width=0.95\columnwidth]{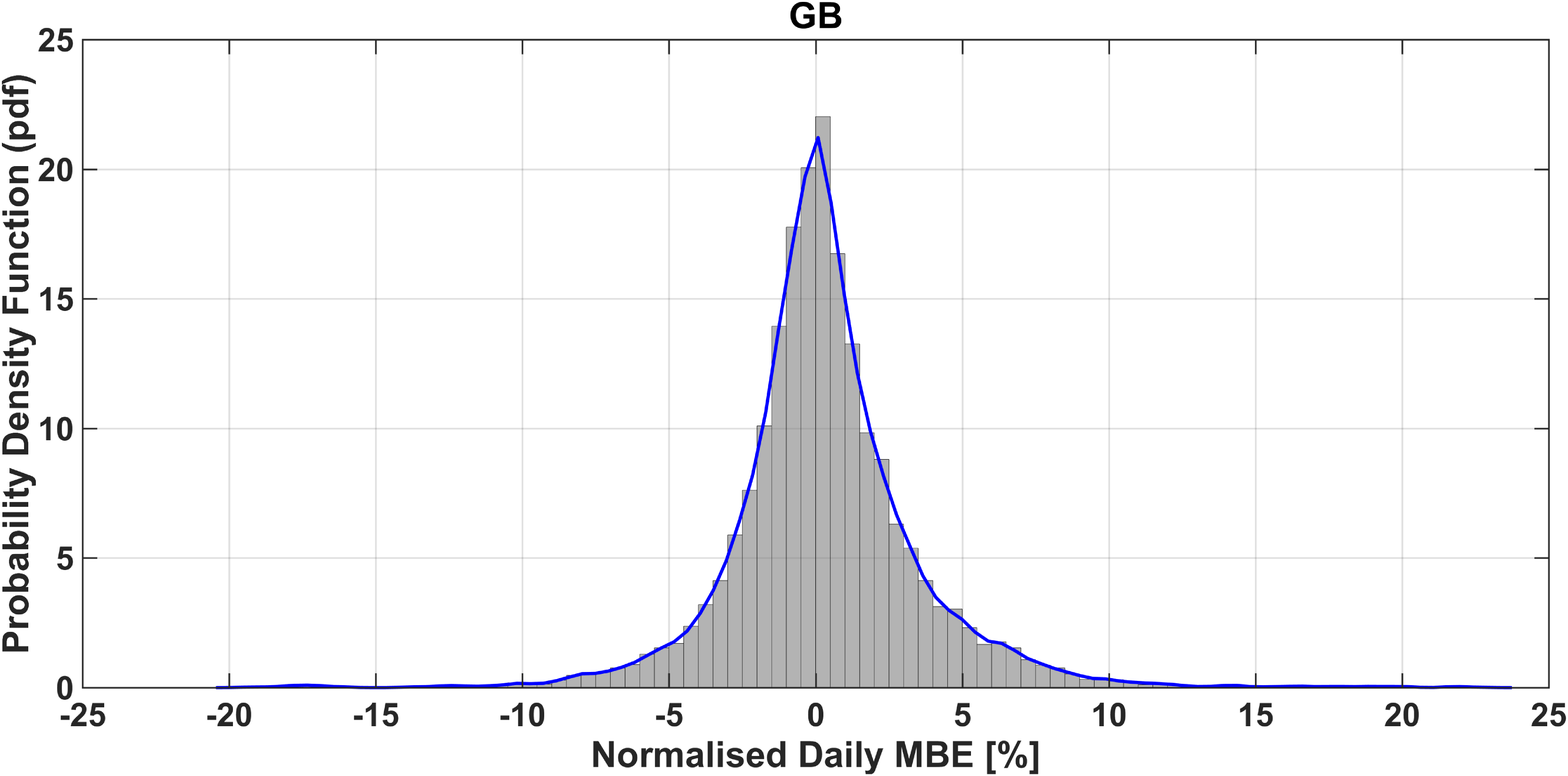} &
    \includegraphics[width=0.95\columnwidth]{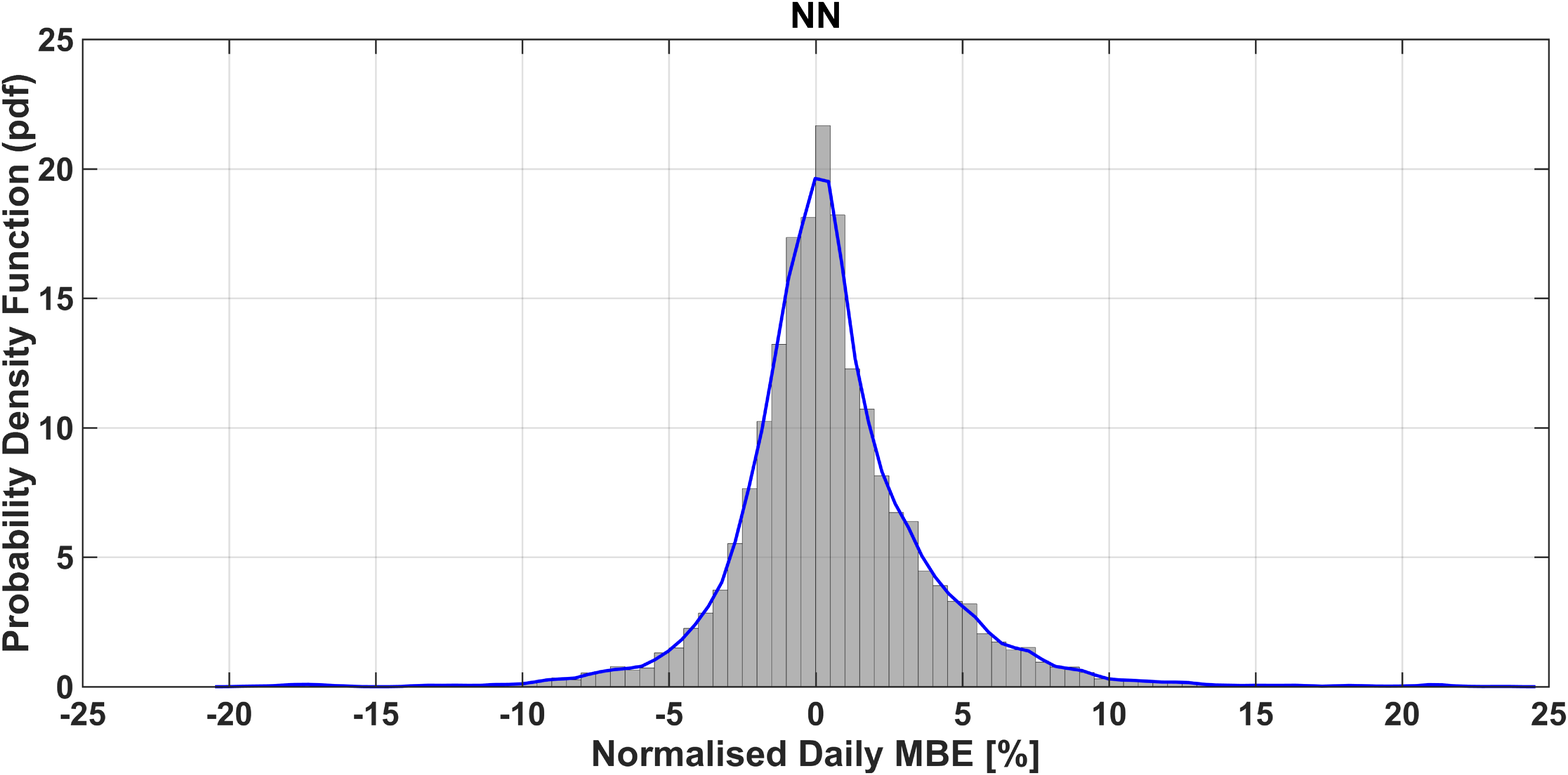} \\  
    \includegraphics[width=0.95\columnwidth]{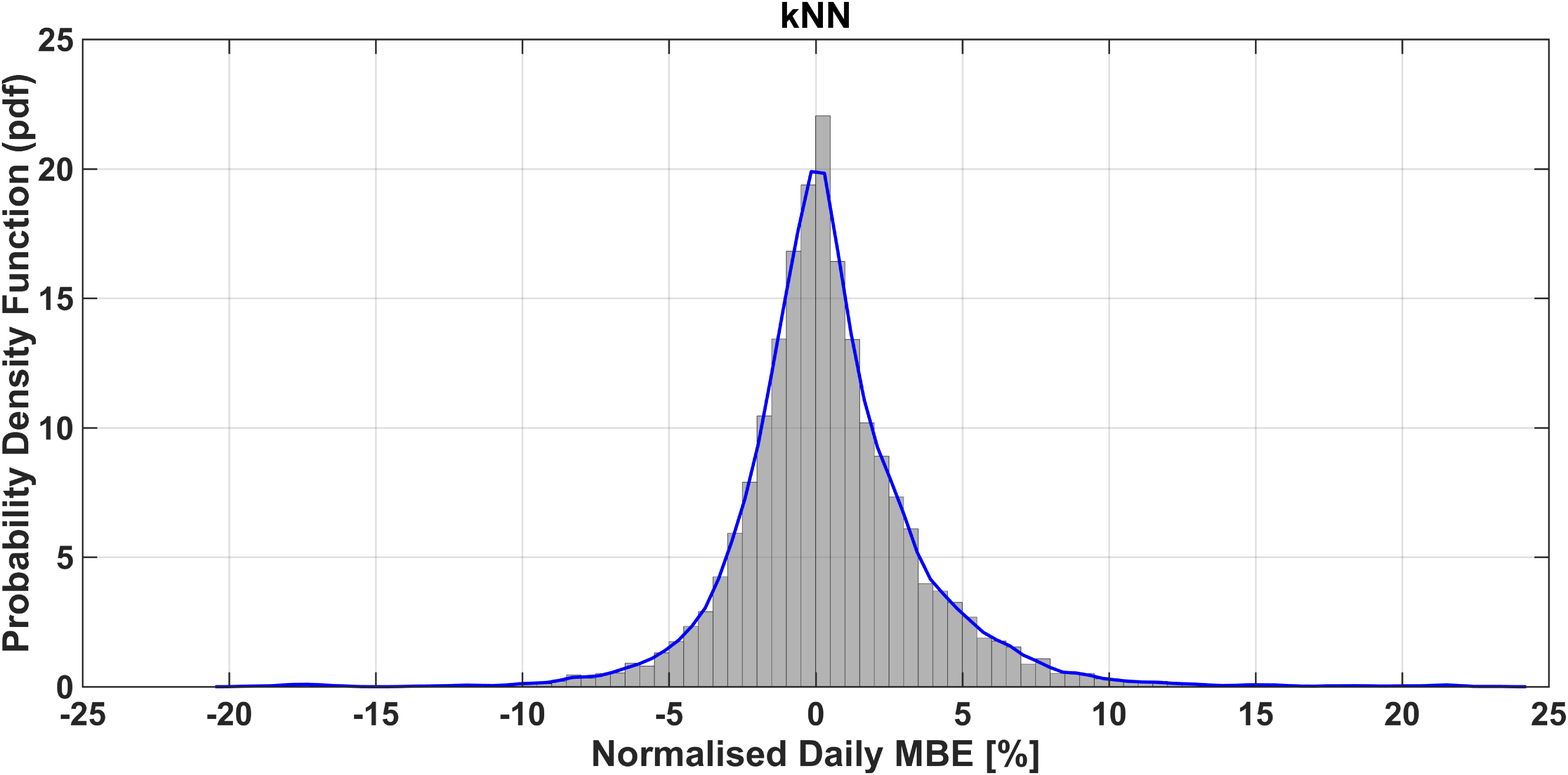} &
    \includegraphics[width=0.95\columnwidth]{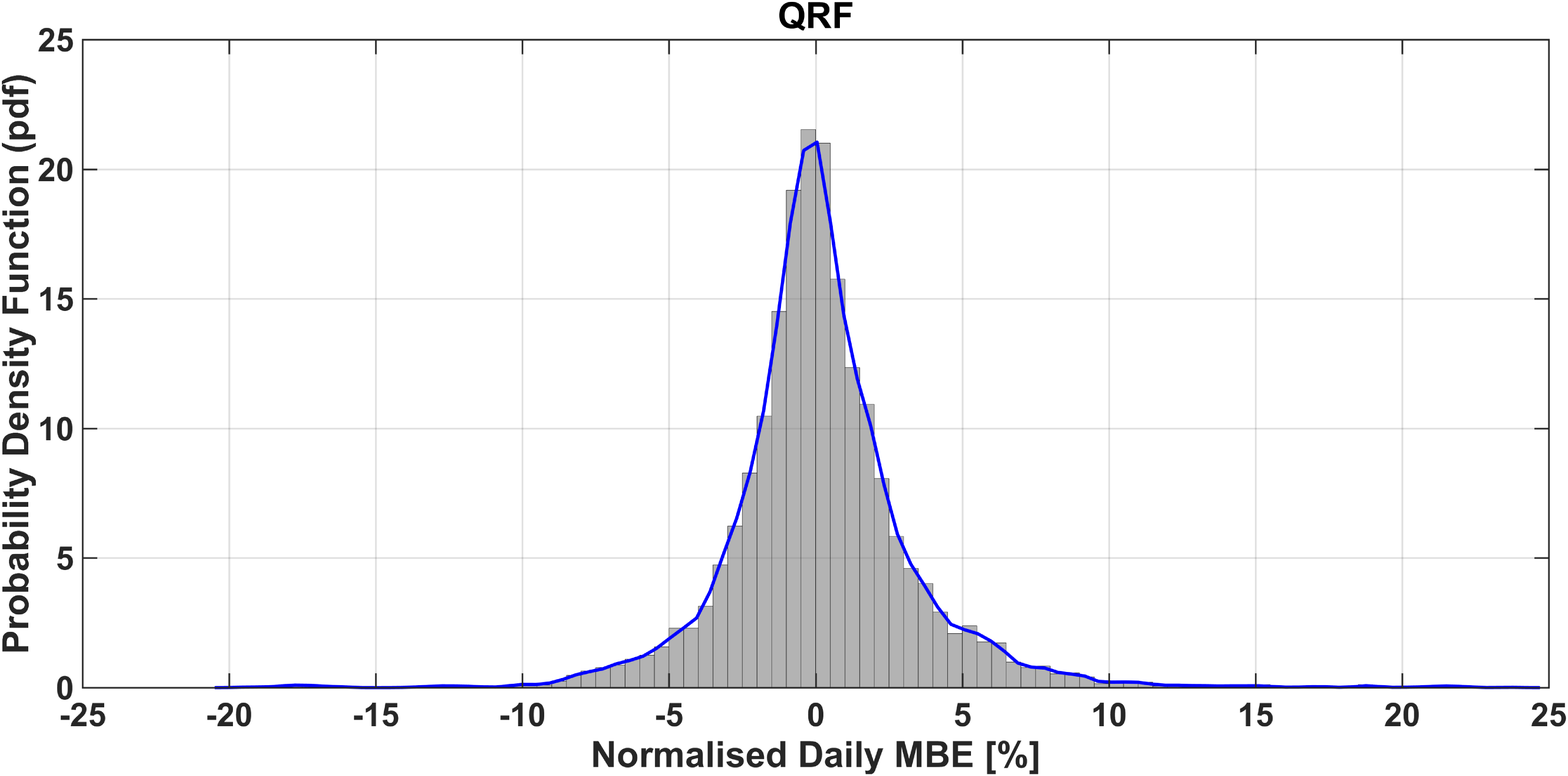} \\
		\includegraphics[width=0.95\columnwidth]{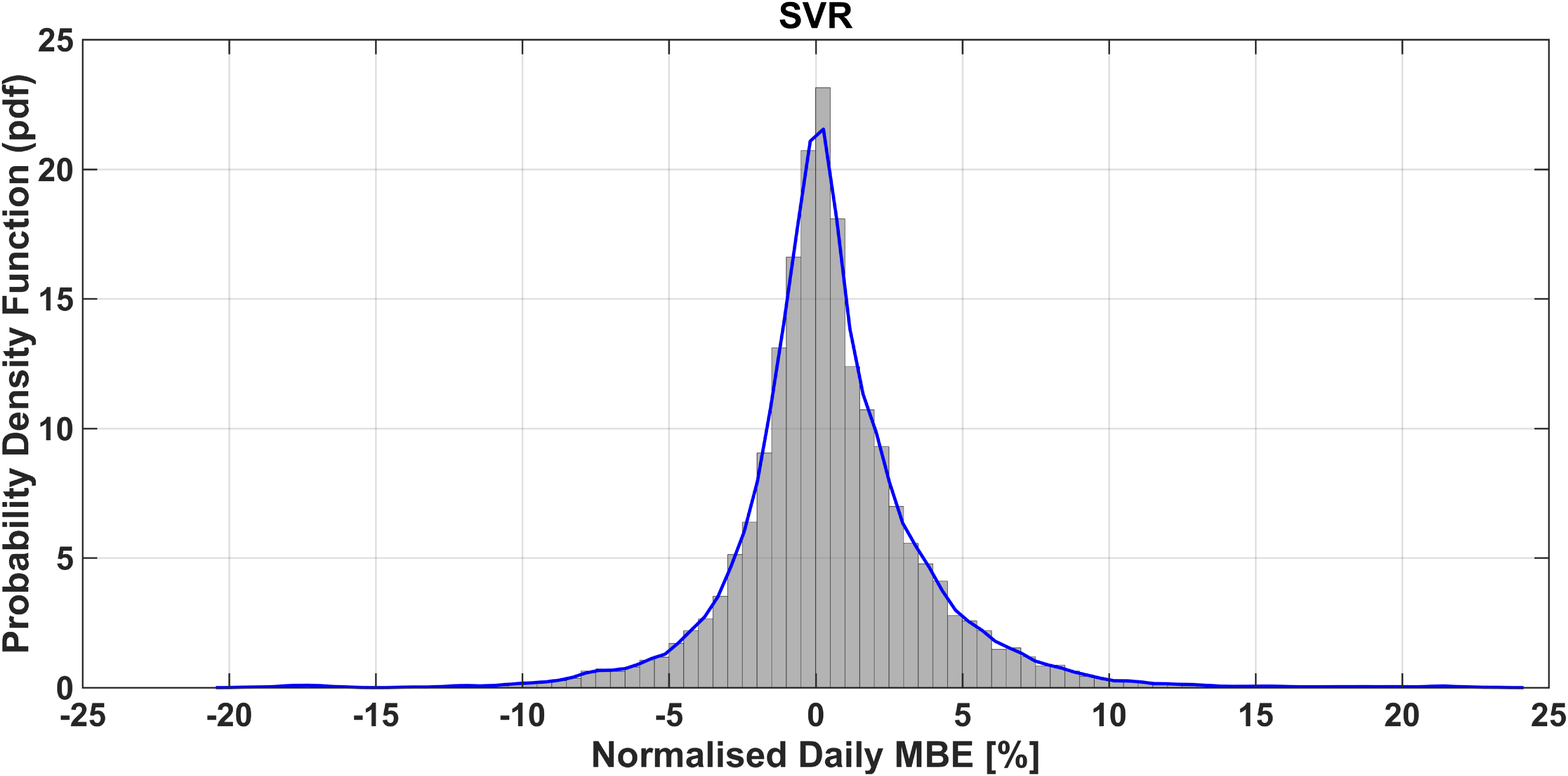} &
    \includegraphics[width=0.95\columnwidth]{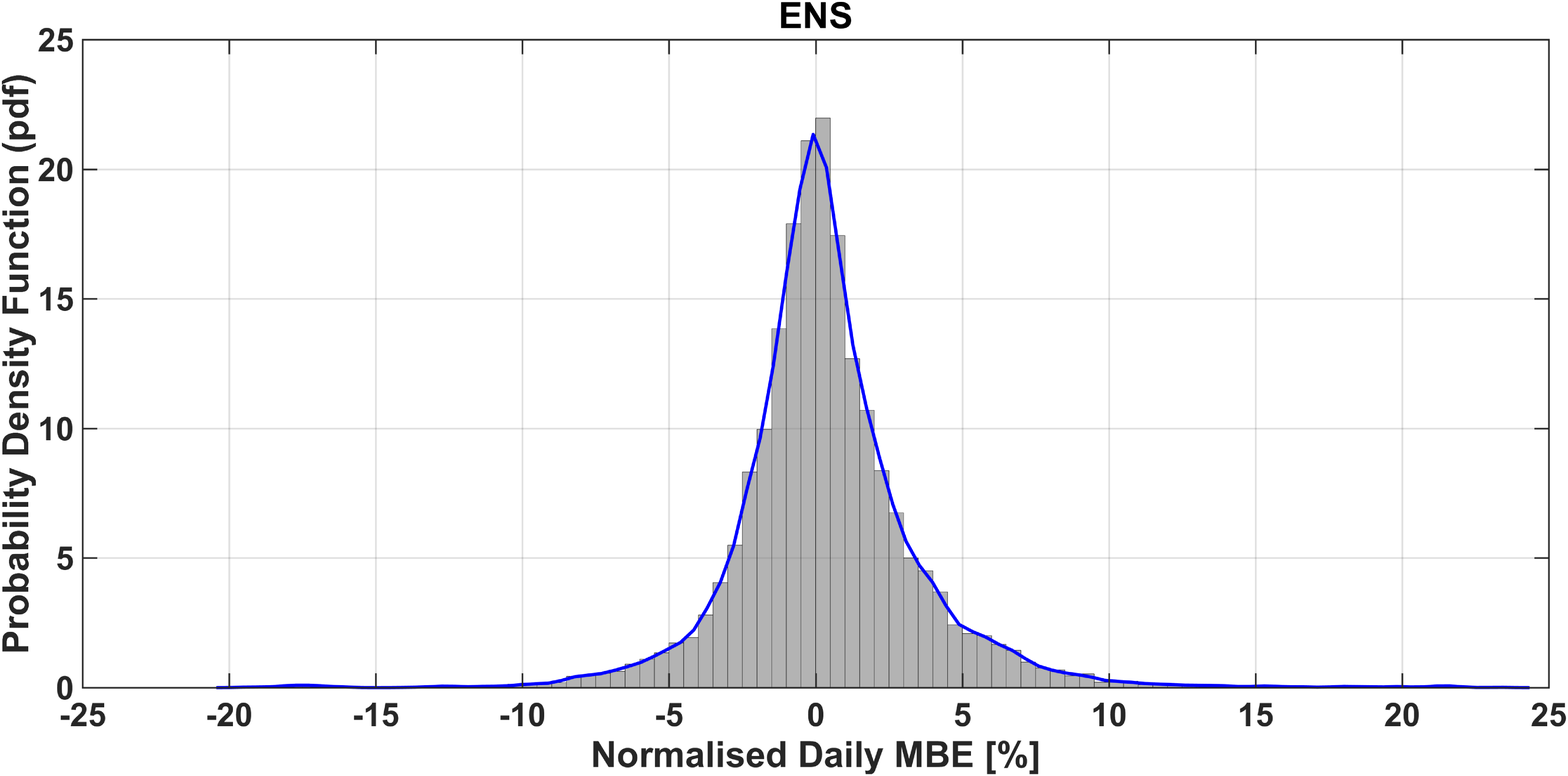} \\
		  \end{tabular}
		  }
  \caption{Histogram of the occurrences of given daily nMBE errors, and plot of the corresponding Probability Density Functions (PDFs) for the six forecasting methodologies.}
  \label{NMBE_Histogram}
  \vspace{-1mm}
\end{figure*}
The full normalized histogram of the (average) daily nMBE error is also shown in Figure \ref{NMBE_Histogram} for each one of the six methods, together with the corresponding (truncated) kernel-smoothing estimation of the probability density function (PDF) (see \cite{Bowman1997}). From this figure, it can be also noticed that there are some days when the nMBE is very large (bigger than 20\%). We also noticed that large occurrences of the nMBE happen at the same time for all methods. This could thus be caused by a completely wrong weather forecast or, more likely, to a wrong measurement of generated power, or to other unknown failures (or not communicated maintenance works) that occurred in the monitored PV plants.\\
\newline
Finally, from the histograms it is possible to appreciate again that in our analysis ENS outperforms the other algorithms. In fact, the lowest three values of the variance of the PDFs were obtained by the ENS, QRF and kNN methods respectively (ENS:$9.93$, QRF:$10.02$,  kNN:$10.13$). Given the (relatively) small improvement, we validated this result using the signed rank Wilcoxon test \cite{Wilcoxon}, considering a significance level of 5\%. The test can be used to compare two algorithms at the same time, and we used it to assess the statistical significance of the nMAE difference, by using all the hourly error values. As a result, after comparing every possible pair of forecasting algorithms, we obtained that all the differences could be believed as statistically significant, with the highest value (i.e., smallest difference) obtained when comparing kNN and QRF (p-value equal to 0.019). Even in such case however the values are still well below the 5\% threshold, and the differences may still be regarded as significant.

\subsection{Dependence on weather conditions}
It is well known that weather conditions have an impact on power generation from PV plants, as well as on the accuracy of its prediction (see for instance \cite{Yang2014}). In this section we shall both evaluate how the accuracy of the predictions changes under different weather conditions, as well as what algorithms are most performing in such circumstances. For this purpose, we shall use the \textit{clear sky index} (CSI) to distinguish different weather conditions. The clear sky index is defined as the ratio between the actual global insolation measured at the site, and the global insolation expected if the sky were clear \cite{Mills2011}. Many authors have used the CSI for this purpose, and for instance the authors of \cite{Lorenz2009} have noticed that there may be a considerable overestimation of the irradiance for cloudy situations with CSI index between 0.3 and 0.8, while the actual measured irradiance may be underestimated when the CSI is lower than 0.2.\\
\newline
Table \ref{CSI_nMAE} reports the average hourly nMAE error obtained in all PV plants by each method for different values of the CSI.
\begin{table}[tb!]
	\caption{Average nMAE error under different weather conditions.} \centering
	\resizebox{\linewidth}{!}{
	\begin{tabulary}{\columnwidth}{|l||l||l|l|l|l|l|l|}
		\hline
		CSI / Method & Number of hours & GB & NN & kNN & QRF & SVR & ENS\\
		\hline
		\hline
		CSI $\leq 0.1$ & 553 & 6.61 & 6.75 & 6.64 & 6.63 & 6.71 & \textbf{6.61}\\
		\hline
		$0.1 <$ CSI $\leq 0.2$ & 3600 & \textbf{12.5} & 13.3 & 13.8 & 13.1 & 13.0 & 13.1\\
		\hline
		$0.2 <$ CSI $\leq 0.3$ & 9377 & 8.38 & 8.82 & 8.91 & \textbf{8.31} & 8.57 & 8.45\\
		\hline
		$0.3 <$ CSI $\leq 0.4$ & 9311 & 6.90 & 7.27 & 6.95 & 6.51 & 7.01 & \textbf{6.69}\\
		\hline
		$0.4 <$ CSI $\leq 0.5$ & 7055 & 7.07 & 7.39 & 6.86 & \textbf{6.53} & 7.13 & 6.72\\
		\hline
		$0.5 <$ CSI $\leq 0.6$ & 6200 & 6.77 & 6.98 & 6.42 & \textbf{6.25} & 6.84 & 6.37\\
		\hline
		$0.6 <$ CSI $\leq 0.7$ & 7122 & 6.97 & 7.00 & 6.23 & \textbf{6.21} & 6.75 & 6.29\\
		\hline
		$0.7 <$ CSI $\leq 0.8$ & 8366 & 6.63 & 6.54 & \textbf{5.94} & 6.08 & 6.46 & 6.00\\
		\hline
		$0.8 <$ CSI $\leq 0.9$ & 13212 & 5.84 & 5.59 & \textbf{5.08} & 5.33 & 5.56 & 5.16\\
		\hline
		$0.9 <$ CSI $\leq 1$ & 57324 & 6.38 & 6.11 & 6.06 & 6.16 & 6.07 & \textbf{5.94}\\
 	\hline
	\end{tabulary}
	}
	\label{CSI_nMAE}
\end{table}
As can be seen in Table \ref{CSI_nMAE}, predictions are less accurate for values of the CSI between 0.1 and 0.2, while they are more accurate for values greater than 0.8. This result is consistent with what had been observed in \cite{Lorenz2009}. In comparative terms, it is also possible to see that the GB method is the most performing one when the sky is cloudy, QRF provides the best performance for intermediate values of the CSI, while the kNN is the best methodology when the sky is clear. Overall, the GB is clearly penalised by the fact that in Italy high values of CSI occur most frequently (see the second column of Table \ref{CSI_nMAE}). In this framework, the ensemble is still the most valuable approach as it always provides one of the best forecasts under any weather condition.\\
\newline
\textbf{Remark: }note that the CSI is not well defined for night hours, when the expected insolation at the denominator is zero. Accordingly, night hours were not considered in Table \ref{CSI_nMAE}, and this gave rise to larger nMAE errors than in the previous comparisons and tables.

\subsection{Importance of accurate weather forecasts}

\begin{figure*}[!t]
\centering
\includegraphics[width=\columnwidth]{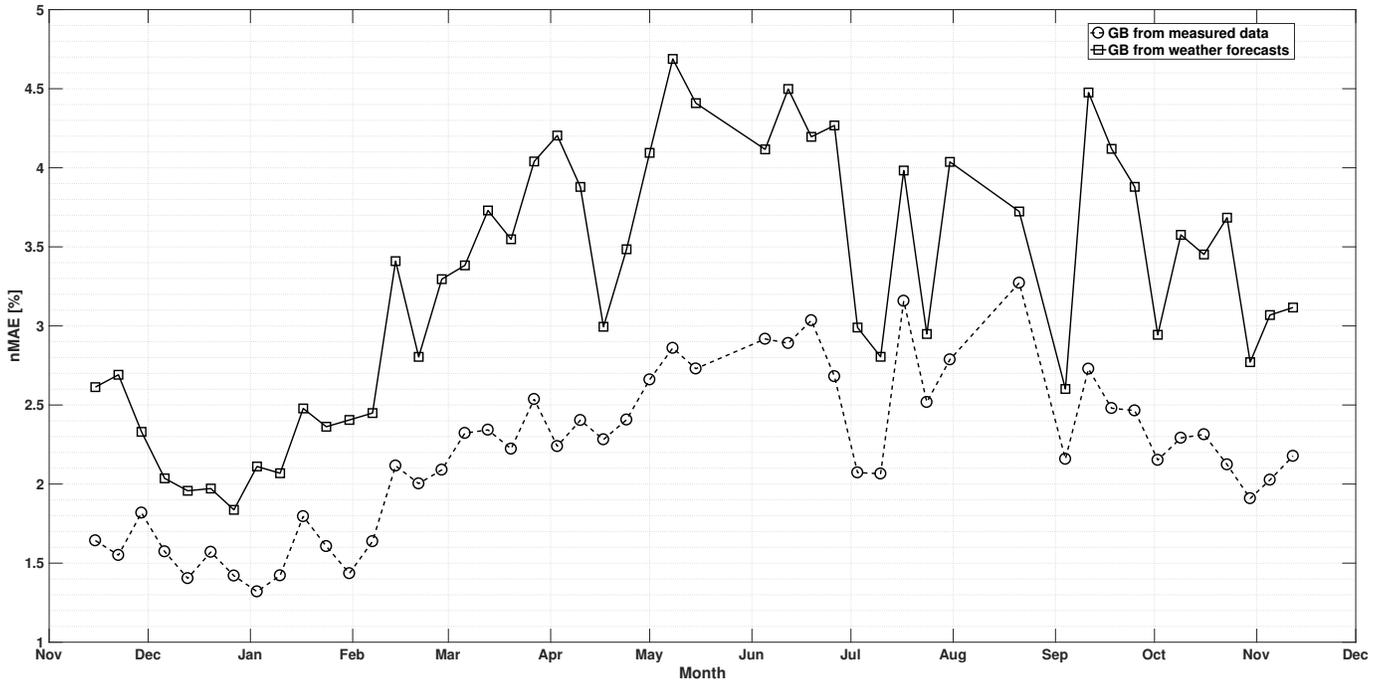}
\caption{The use of more accurate weather forecasts of the global tilted irradiance may provide an improvement of about 1\% to the GB methodology.}
\label{GB_with_real_data}
\end{figure*}

The accuracy of the prediction of the expected energy produced by PV plants heavily depends, among others, on the accuracy of available meteorological forecasts. Here, we are interested in evaluating the margin of improvement that can be gained by having more accurate (possibly exact) meteorological forecasts. In particular, we now use the available irradiance measured from satellite data (with resolution of 3.5 km $\times$ 3.5 km) as an alternative to the GTI forecasts as an input to the GB method. The satellite data that we had at our disposal had a time resolution of 1 hour (and provided data aggregated with a sampling time of 15 minutes). Results are illustrated in Figure \ref{GB_with_real_data} and show that an improvement of about 1\% can be consistently obtained over the whole year.
Although satellite data are clearly affected by some measurement errors (e.g., due to the resolution of the data), still they show that a significant improvement can be obtained with respect to the use of meteorological forecasts.\\
\newline
\textbf{Remark: }we noticed that similar improvements are obtained with the other methods as well, by substituting the predicted GTI with the measured one (with the ultimate average error of ENS around 2.00\%). However, we have only reported the results for the GB, as in this case we could directly substitute the predicted GTI with the measured one, and we could directly evaluate the impact of (not perfect) weather forecasts. On the other hand, in the case of QRF for instance, we did not have the measurements of some input variables (e.g., BTI or DTI) to evaluate the same impact.

\section{Conclusion}
\label{Conclusion}
Despite the rich literature on forecasting techniques for PV plants, the lack of thorough comparisons among different techniques in (spatially and temporally) extended datasets motivated the present work. Among the main results of this paper, we have evaluated the difference in the accuracy of simple methodologies (GB method) and more complicated approaches (QRF method, and ensemble of methods). While we noticed that the difference is not too great (an overall improvement in the nMAE of about 5\%), still the improvement was shown to be consistent, and statistically relevant, over all PV plants and months of a year. Different methodologies appear to be the most suitable under different weather conditions (GB providing the best results in cloudy conditions, QRF for intermediate values of the CSI, and the kNN in sunny conditions). In this context, the ensemble has the advantage of providing one of the best forecasts under any weather condition. We also showed that more accurate weather forecasts of the irradiance alone would already improve the accuracy from (about) 3\% to 2\%.\\
\newline
In our experience, the residual $2\%$ error is mainly due to the fact that measured data (and forecasts as well) had a time resolution of 1 hour. Accordingly, they may fail to notice some effects occurring at a much faster time scale. Other sources of errors are that the irradiance (and its beam and diffuse components) are not obviously exactly the same everywhere in the PV plant; a PV plant contains a number of mechanical/electrical elements that in real life may give rise to some unexpected behaviours; and, finally, we are aware that there may have been some measurement errors, or some approximations, or even some transcription/registration errors in the available datasets.\\
\newline
As a final remark, in our experience we noticed that all methodologies performed in a better way if temperature data were not considered. Given that it is well-known that temperature is a variable that does contribute (though through higher-order terms) to the actual energy generation, our intuition is that the predicted temperatures that we had at our disposal were not accurate enough to in fact improve our forecasts. To validate this hypothesis we have started collecting temperature measurements to evaluate the ultimate improvement of having this further variable, at least in the ideal case where meteorological variables are measured. This will allow us to assess the sensitivity of each method to this specific input parameter.


\begin{thebibliography}{10}

\bibitem{Marinelli2014}
	M.~Marinelli, F.~Sossan, G.T.~Costanzo, and H.W.~Bindner, \emph{Testing of a Predictive Control Strategy for Balancing Renewable Sources in a {Microgrid}}, IEEE Transactions on Sustainable Energy, Vol.~5, No.~4, pp.~1426-1433, 2014.

\bibitem{Tucci2016}
	M.~Tucci, E.~Crisostomi, G.~Giunta, and M.~Raugi, \emph{A Multi-Objective Method for Short-Term Load Forecasting in {European} Countries}, IEEE Transactions on Power Systems, Vol.~6, No.~1, pp.~104-112, 2016.

\bibitem{Yona2013}
	A.~Yona, T.~Senjyu, T.~Funabashi, and C.-H.~Kim, \emph{Determination Method of Insolation Prediction With Fuzzy and Applying Neural Network for Long-Term Ahead {PV} Power Output Correction}, IEEE Transactions on Sustainable Energy, Vol.~4, No.~2, pp.~527-533, 2013.
	
\bibitem{Bizzarri2013}
	F.~Bizzarri, M.~Bongiorno, A.~Brambilla, G.~Gruosso, and G.S.~Gajani, \emph{Model of Photovoltaic Power Plants for Performance Analysis and Production Forecast}, IEEE Transactions on Sustainable Energy, Vol.~4, No.~2, pp.~278-285, 2013.
	
\bibitem{Morales2014}
J.M.~Morales, A.J.~Conejo, H.~Madsen, P.~Pinson, and M.~Zugno, \emph{Integrating Renewables in Electricity Markets}, Springer, New York 2014.

\bibitem{Kariniotakis2017}
G.~Kariniotakis, \emph{Renewable Energy Forecasting, From Models to Applications}, 1st Edition, Elsevier, 2017.
	
\bibitem{Yang2015}
	C.~Yang, A.A.~Thatte, and L.~Xie, \emph{Multitime-Scale Data-Driven Spatio-Temporal Forecast of Photovoltaic Generation}, IEEE Transactions on Sustainable Energy, Vol.~6, No.~1, pp.~104-112, 2015.

\bibitem{Shah2015}
	A.S.B.M.~Mohd Shah, H.~Yokoyama, and N.~Kakimoto, \emph{High-Precision Forecasting Model of Solar Irradiance Based on Grid Point Value Data Analysis for an Efficient Photovoltaic System}, IEEE Transactions on Sustainable Energy, Vol.~6, No.~2, pp.~474-481, 2015.

\bibitem{Liu2015}
	J.~Liu, W.~Fang, X.~Zhang, and C.~Yang, \emph{An Improved Photovoltaic Power Forecasting Model With the Assistance of Aerosol Index Data}, IEEE Transactions on Sustainable Energy, Vol.~6, No.~2, pp.~434-442, 2015.
	
\bibitem{Larson2016}
	D.P.~Larson, L.~Nonnenmacher, and C.F.M.~Coimbra, \emph{Day-ahead forecasting of solar power output from photovoltaic plants in the American Southwest}, Renewable Energy, Vol.~91, pp.~11-20, 2016.

\bibitem{Ehsan2016}
	R.~Muhammad Ehsan, S.P.~Simon, and P.R.~Venkateswaran, \emph{Day-ahead forecasting of solar photovoltaic output power using multilayer perceptron}, Springer Neural Computing \& Applications, pp.~1-12, 2016.

\bibitem{Ogliari2016}
	E.~Ogliari, A.~Gandelli, F.~Grimaccia, S.~Leva, and M.~Mussetta, \emph{Neural forecasting of the day-ahead hourly power curve of a photovoltaic plant}, IEEE International Joint Conference on Neural Networks (IJCNN), pp.~1-5, 2016.
	
\bibitem{Yang2014}
	H.-T.~Yang, C.-M.~Huang, Y.-C.~Huang, and Y.-S.~Pai, \emph{A weather-based hybrid method for 1-day ahead hourly forecasting of PV power output}, IEEE Transactions on Sustainable Energy, Vol.~5, No.~3, pp.~917-926, 2014.
	
\bibitem{Li2015}
	Z.~Li, C.~Zang, P.~Zeng, H.~Yu, and H.~Li, \emph{Day-ahead Hourly Photovoltaic Generation Forecasting using Extreme Learning Machine}, 5th Annual IEEE International Conference on Cyber Technology in Automation, Control and Intelligent Systems, Shenyang, China, 2015.
	
\bibitem{Raza2016}
	M.Q.~Raza, M.~Nadarajah, and C.~Ekanayake, \emph{On recent advances in {PV} output power forecast}, Solar Energy, Vol.~136, pp.~125-144, 2016.
	
\bibitem{Antonanzas2016}
	J.~Antonanzas, N.~Osorio, R.~Escobar, R.~Urraca, F.J.~Martinez-de-Pison, and F.~Antonanzas-Torres, \emph{Review of photovoltaic power forecasting}, Solar Energy, Vol.~136, pp.~78-111, 2016.

\bibitem{Hong2016}
	T.~Hong, P.~Pinson, S.~Fan, H.~Zareipour, A.~Troccoli, and R.J.~Hyndman, \emph{Probabilistic energy forecasting: Global Energy Forecasting Competition 2014 and beyond}, International Journal of Forecasting, Vol.~32, pp.~896-913, 2016.
	
\bibitem{Rokach2010}
	L.~Rokach, \emph{Ensemble-based classifiers}, Artificial Intelligence Review, Vol.~33, No.~1-2,pp.~1–39, 2010.
	
\bibitem{Perez1990}	
	R.~Perez, P.~Ineichen, R.~Seals, and R.~Stewart, \emph{Modeling daylight availability and irradiance components from direct and global irradiance}, Solar Energy, Vol.~44, No.~5, pp.~271-289, 1990.
	
\bibitem{DeGiorgi2014}
	M.G.~De Giorgi, P.M.~Congedo, and M.~Malvoni, \emph{Photovoltaic power forecasting using statistical methods: impact of weather data}, IET Science, Measurement and Technology, Vol.~8, No.~3, pp.~90-97, 2014.
	
\bibitem{IEA2012} 
	R.H.~Inman, H.T.C.~Pedro, and C.F.M.~Coimbra, \emph{Solar forecasting methods for renewable energy integration}, Progress in Energy and Combustion Science, Vol.~39, pp.~535-576, 2013.
	
\bibitem{PVUSA}
	R.~Dows, and E.~Gough, \emph{{PVUSA} procurement, acceptance, and rating practices for photovoltaic power plants}, Pacific Gas and Electric Company, San Ramon, CA, Tech. Rep., 1995.

\bibitem{Bianchini2013} 
G.~Bianchini, S.~Paoletti, A.~Vicino, F.~Corti, and F.~Nebiacolombo, \emph{Model estimation of photovoltaic power generation using partial information}, $4^{th}$ IEEE PES Innovative Smart Grid Technologies Europe (ISGT Europe) Conference, Copenhagen, Denmark, pp.~1-5, 2013.

\bibitem{Breiman2001}
L.~Breiman, \emph{Random Forests}, Machine Learning, Vol.~45, No.~1, pp.~5-32, 2001.

\bibitem{Meinshausen2006}
M.~Meinshausen, \emph{Quantile Regression Forests}, Journal of Machine Learning Research, Vol.~7, pp.~983-999, 2006.

\bibitem{Vapkin1995}
C.~Cortes, and V.~Vapnik, \emph{Support-vector networks}, Machine learning, Vol.~20, No.~3, pp.~1273-1297, 1995.

\bibitem{Vapkin1997}
V.~Vapkin, S.E.~Golowich, and A.J.~Smola, \emph{Support vector method for function approximation, regression estimation and signal processing}, In: Advances in neural information processing systems, ed. by M.~Mozer and M.~Jordan and T.~Petsche, pp.~281–287, Cambridge, MA, MIT Press, 1997.

\bibitem{Smola2004}
A.J.~Smola, and B.~Sch\"{o}lkopf, \emph{A tutorial on support vector regression}, Statistics and computing, Vol.~4, No.~3, pp.~199-222, 2004.

\bibitem{Wolpert1992}
Wolpert, D., \emph{Stacked Generalization}, Neural Networks, Vol.~5, No.~2, pp.~241-259, 1992.

\bibitem{Alessandrini2015}
S.~Alessandrini, L.~Delle Monache, S.~Sperati, and G.~Cervone  \emph{An analog ensemble for short-term probabilistic solar power forecast}, Applied Energy, Vol.~157, pp.~95-110, 2015.

\bibitem{Ren2015}
Y.~Ren, P.N.~Suganthan, and N.~Srikanth, \emph{Ensemble methods for wind and solar power forecasting — A state-of-the-art review}, Renewable and Sustainable Energy Reviews, Vol.~50, pp.~82-91, 2015.

\bibitem{Bowman1997}
A.W.~Bowman, and A.~Azzalini, \emph{Applied Smoothing Techniques for Data Analysis}, New York: Oxford University Press Inc., 1997.

\bibitem{Wilcoxon}
N.S.~Siegel, and J.~Castellan~Jr., \emph{Non parametric Statistics for the Behavioral Sciences}, 2nd Edition, McGraw-Hill, New York, NY, USA, 1988.

\bibitem{Mills2011}
A.D.~Mills, and R.H.~Wiser, \emph{Implications of geographic diversity for short-term variability and predictability of solar power}, IEEE PES General Meeting, Detroit, MI, USA, pp.~1-9, 2011.

\bibitem{Lorenz2009}
E.~Lorenz, J.~Hurka, D.~Heinemann, and H.G.~Beyer, \emph{Irradiance Forecasting for the Power Prediction of Grid-Connected Photovoltaic Systems}, IEEE Journal of Selected Topics in Applied Earth Observations and Remote Sensing, Vol.~2, No.~1, pp.~2-10, 2009.



\end{thebibliography}
\end{document}